# Use Pronunciation by Analogy for text to speech system in Persian Language

Ali Jowharpour[1], DR. Masha allah Abbasi dezfuli[1], DR. Mohammad Hosein Yektaee[1]

[1]Islamic Azad university Science and research branch-khuzestan Iran

**Abstract**
The interest in text to speech synthesis increased in the world . text to speech have been developed for many popular languages such as English, Spanish and French and many researches and developments have been applied to those languages. Persian on the other hand, has been given little attention compared to other languages of similar importance and the research in Persian is still in its infancy. Persian language possess many difficulty and exceptions that increase complexity of text to speech systems. For example: short vowels is absent in written text or existence of homograph words. in this paper we propose a new method for persian text to phonetic that base on pronunciations by analogy in words, semantic relations and grammatical rules for finding proper phonetic.

## 1.Introduction

Many text-to-speech (TTS) systems use look-up in a large dictionary as the primary strategy to determine the pronunciation of input words. However, it is not possible to list exhaustively all the words of a language, so that a secondary or 'back-up' strategy is required for words not in the system dictionary. Pronunciation by analogy (PbA) is a data-driven technique for the automatic phonemization of text, first proposed over a decade ago by Dedina and Nusbaum [1,2]. Although initially PbA attracted little attention, several groups around the world are now trying to develop the approach. There is accumulating evidence that PbA easily outperforms linguistic rewrite rules [3, 4, 5, 6] .In this paper, we extend previous work on PbA in directions which are intended to improve its usability within the pronunciation component of a TTS system. We have studied extended methods both for pattern matching (between the input word and the dictionary entries) . This has produced improvements on the best results so far reported in the persian language. but these improvements have so far been only small. this paper compares different methods for persian TTS. In this paper describe detecting of word's phonetic by similarity of letters in words and explain this method in part 3.

## 2.Related Works

In PbA, an unknown word is pronounced by matching substrings of the input to substrings of known, lexical words, hypothesizing a partial pronunciation for each matched substring from the phonological knowledge, and assembling the partial pronunciations. Here, we use an extended and improved version of the system described by Dedina and Nusbaum (1991), which consists of four components: the (uncompressed and previously aligned) lexical database, the matcher which compares the target input to all the words in the database, the pronunciation lattice (a data structure representing possible pronunciations), and the decision function, which selects the 'best' pronunciation among the set of possible ones. The lexicon used is *Webster's Pocket Dictionary,* containing 20,009 words manually aligned by Sejnowski and Rosenberg (1987) for training their NETtalk neural network[7].

The other work is done by Namnabat and Homayounpour in Amirkabir University of Technology. They have constructed a system including a rule based section and multi layer perceptron (MLP) neural network and the ultimate accuracy of their system is 87% (Namnabat and Homayounpour,2006).[8].

## 3. pronunciation of input word

Figure 1 shows a block diagram of the pronunciation system that will explain in this paper.  Prepare the code for input word and  compare it  with sample words in database then select matching  words  and given to next step in block diagram. recognition network block identify most similar sample to input word and added short vowels to input word. subsequently changed input word send to speech phase. Note that there is no guarantee that there will always be complete phonetic for every input word.

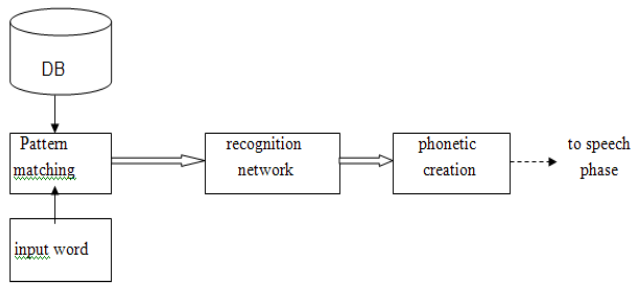

Fig.1. block diagram of the pronunciation system

### 3.1 Input word coding

When word give to system, for simplify replace its letters by number. These numbers shows in table 1. coding divided words to different groups base on structure similarity.

Table 1. code for letters

| consonants | 7 |
|---|---|
| Long vowel "ا"    "ā" | 1 |
| Long vowel "ی"    "i" | 2 |
| Long vowel "و"    "u" | 3 |

### 3.2 Pattern matching

pattern matching process starts with making suitable code for input string. Rules of code making is 7 for consonant and 1,2,3 for long vowels base on table 1. Then made code compare with words' code in DB. Then selected words in DB that have the same input word's code.

### 3.3 Pronunciation dictionaries (DB)

The re-sources include a pronunciation dictionary for persian language with about 2000 entries, which were used in the persian TTS. Advantage of this method, using small database that contain patterns, word's code, phonetic and grammatical kind of each entry. Selection entries in database based on coverage uttermost input words in same length and trait.

### 3.4 Recognition Network

At this stage the most similar word to input word will be found. Fig 2 shows the main form of recognition network block.

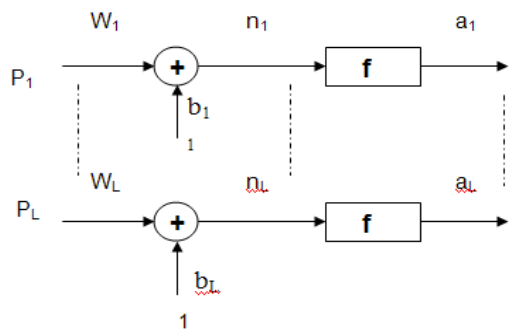

Fig 2.recognition network.

As can be seen in fig 2. $P_i$ is ascii code of pattern letters that is considered input for each neuron. $W_i$ is coefficient for all ascii codes that its value is one. $b_i$ value is negative of ascii code for any letters of input word. n is integer for each letter comparison. If $w_i=b_i$ then n=0 else n<>0. F is function that return $[e^{-|n|}]$.

This function operate similar impulse function. With zero input, the output is 1 and else the output is zero. Thus for similar letters output is one and else is zero. All $a_i$ s make the vector that number of elements equals to input word length and the values of their elements are zero or one. This vector is considered as network output. It is clear that output of this network layer is matrix that number of rows are number of selected pattern and number of columns are length of input word.

After creating output matrix create a vector that each element is letter influence proportion in word structure. Long vowels have the most role in word pronunciation and consonants less. therefore consonant weight is consider one and vowel weight is length of the input word. If $\omega p_i$ be i th letter weight and L be length of the input word:

$$\omega_i = \begin{cases} 1 & i\ th\ letter\ is\ consonant \\ L & i\ th\ letter\ is\ long\ vowel \end{cases}$$

$\omega = \sum_{i=1}^{l} \omega_i a_i$     $\omega$ is total weight of word

for finding most similar pattern to input word by calculate $s=\omega p/\sum_{i=1}^{l} \omega i_i$ that $\omega p$ is weight of pattern and $\omega i_i$ is ith letter weight of input word. Result is value between 0 and 1. If result be near 1 mean is pattern is more similar to input word. When found similarity of all pattern words to input word then sort them and select pattern that have maximum s. if s value of first pattern is equal to s value of next pattern may there is homograph words. If s value of second word is one; one of the samples must be selected according to sentence concept If possible else use word iteration frequency in the persian text. For better explanation, see the example.

Example: assume input word is "رنگ" ("r","n","g") and 777 generated for it. Each word with 777 code select From DB, such as "ونک" (vanak),"ترک" (tork), "سنگ" (sang) and "خرس" (xers) give to recognition network. Network makes 4*3 matrix that each its row is letter's weight of one pattern and a 3 element vector that each its element is weight of "ر","ن","گ" letters.

$$\begin{bmatrix} 0 & 1 & 0 \\ 0 & 0 & 0 \\ 0 & 1 & 1 \\ 0 & 0 & 0 \end{bmatrix} \times \begin{bmatrix} 1 \\ 1 \\ 1 \end{bmatrix} = \begin{bmatrix} 1 \\ 0 \\ 2 \\ 0 \end{bmatrix}$$

Because there is not long vowel in "رنگ" all elements of vector are one.

Each element in result vector is $\omega p$ for one pattern and $\omega i=1+1+1=3$.

S values are: S("ونک")=.333, S("ترک")=0, S("سنگ")=.666 and S("خرس")=0 therefore suitable pattern is "سنگ".

### 3.5 phonetic creation

In previous example selected pattern was "سنگ" with "sang" phonetic. At this case use short vowel that is /a/ in "sang", and insert it between "r" and "n" in "r n g" and complete phonetic (rang).

## 4. Evaluation

The evaluation of the system was performed using two different corpuses:

The first is a list of word (about 2000 word) that some of are root of words that constitute patterns in DB. One of avantage of method using of little pattern. This property can use, to apply this method on mobile phone.

The second is accuracy in detecting pronunciation input word by little sample in dictionary.

In tables 2 and 3 the comparison of several TTS systems are shown, the method of this paper is named FPbA.

Table 3. prepare with persian text contain 500 words that if change the text maybe percentage is changed.

Table 2. compare number of pattern in deferent systems

| Text to speech system(title) | Number of pattern in database |
|---|---|
| Multilayer perceptron nueral network.[8] | 98000 |
| Gooya(percian tts) | 45000 |
| Pba.[3] | 50000 |
| Filibuster Swedish[9] | 118104 |
| Filibuster Norwegian[9] | 132806 |
| FPBA | 2000 |

Table 3. accuracy percentage of systems to find phonetic.

| TTS system name | Letter | Word | Sentence |
|---|---|---|---|
| Letter to sound by mlp neural network.[8] | %88 | %61 | ------------ |
| gooya | %91 | %76 | %80 |
| FPBA | %94 | %84 | %69 |

## 5. Conclusion

In this paper we have extended previous works in persian language in several directions. Principally, experimented with a range of pattern matching strategies. In the near future, we intend to use multiple strategies for producing pronunciations. We will use better techniques and better pattern to derive an overall pronunciation. The hope is that, by using the semantic and more grammatical rule, we can produce better pronunciations than any single technique. We should work towards a proper probabilistic model. We have also attempted to produce stress patterns for input words, Finally, we have analyzed common errors of pronunciation.

**Ali jowharpour** received a B.Sc degree of computer hardware from Shiraz university in 1994 and m.sc degree in software from Islamic Azad university Science and research branch-khuzestan Iran in 2011.his research interest is speech technology and signal processing.
**Masha allah Abbasi dezfuli is professor of** Islamic Azad university Science and research branch-khuzestan Iran. his research interest is signal processing.
**Mohammad Hosein Yektaee** is professor of Islamic Azad university Science and research branch-khuzestan Iran. his research interest is multilingual processing.